\documentclass{article} 
\pdfoutput=1
\usepackage{nips15submit_e,times}
\usepackage{hyperref}

\usepackage{url}
\usepackage{import} 
\usepackage{amsmath} 
\usepackage{amssymb}
\usepackage{pdfpages}
\usepackage{graphicx}

\usepackage{algorithm}
\usepackage{algorithmic}

\usepackage{multirow}
\usepackage{wrapfig}
\usepackage{capt-of}

\usepackage[numbers,sort&compress]{natbib}
\usepackage{graphicx} 

\title{Transductive Optimization of Top $k$ Precision}

\author{
Li-Ping Liu ~~~ Thomas G. Dietterich\\
EECS, Oregon State University\\
Corvallis, OR 97330, USA \\
\texttt{\{liuli@eecs.oregonstate.edu, tgd@oregonstate.edu \}} \\
\And
Nan Li ~~~ Zhi-Hua Zhou\\
Department of Computer Science \& Technology, Nanjing University \\
Nanjing 210023, China \\
\texttt{\{lin, zhouzh\}@lamda.nju.edu.cn} \\
}
\newtheorem{theorem}{Theorem}

\newcommand{\bx}{\mathbf{x}}
\newcommand{\by}{\mathbf{y}}

\newcommand{\hbx}{\hat{\mathbf{x}}}
\newcommand{\hby}{\hat{\mathbf{y}}}

\newcommand{\hx}{\hat{x}}
\newcommand{\hy}{\hat{y}}

\newcommand{\indic}{\mathcal{I}}

\nipsfinalcopy 

\begin{document}
\maketitle

\begin{abstract} 
Consider a binary classification problem in which the learner is given a labeled training set, an unlabeled test set, and is restricted to choosing exactly $k$ test points to output as positive predictions.  Problems of this kind---{\it transductive precision@$k$}---arise in information retrieval, digital advertising, and reserve design for endangered species. Previous methods separate the training of the model from its use in scoring the test points.  This paper introduces a new approach, Transductive Top K (TTK), that seeks to minimize the hinge loss over all training instances under the constraint that exactly $k$ test instances are predicted as positive. The paper presents two optimization methods for this challenging problem. Experiments and analysis confirm the importance of incorporating the knowledge of $k$ into the learning process. Experimental evaluations of the TTK approach show that the performance of TTK matches or exceeds existing state-of-the-art methods on 7 UCI datasets and 3 reserve design problem instances.

\end{abstract} 

\section{Introduction}

In the Transductive Precision@$k$ problem, the training set and the unlabeled test set are given, and the task is to predict exactly $k$ test instances as positives. The precision of these selected instances---the fraction of correct positive predictions---is the only measure of importance. Our work is motivated by the problem of designing conservation reserves for an endangered species. Suppose a geographical region is divided into equal-sized cells of land. The species is present in positive cells and absent in negative cells. To protect the species, we seek to purchase some cells (``a conservation reserve''), and we want as many of those as possible to be positive cells.  Suppose we have conducted a field survey of publicly-owned land to collect a training set of cells. With a fixed budget sufficient to purchase $k$ cells, we want to decide which $k$ privately-owned (and un-surveyed) cells to buy. In this paper, we assume that all cells have the same price. This is an instance of the {\it Transductive Precision@$k$} problem.  Other instances arise in information retrieval and digital advertising.

The standard approach to this problem is to first train a classifier on the training data and then threshold the predicted test set scores to obtain the $k$ highest-scoring test instances. Any binary classification algorithm that outputs continuous scores (e.g., an SVM) can be employed in this two-step process. Better results can often be obtained by bipartite ranking algorithms \citep{burges05, rudin09, usunier09, agarwal11, rakotomamonjy12, li14}, which seek to minimize a ranking loss (including ranking losses that put more weight on highly-ranked instances).  In addition to precision@$k$, evaluation measures such as discounted cumulative gain (DCG), normalized discounted cumulative gain (NDCG), and average precision (AP) are often employed to train and evaluate these models. 

Recent work focuses even more tightly on the top-ranked instances. The Accuracy At The Top (AATP) algorithm \citep{boyd12} seeks to optimize the ranking quality for a specified top quantile of the training data.  Maximizing accuracy on the top quantile is intractable, so AATP optimizes a relaxation of the original objective.  

Unlike the ranking problems discussed so far, our problem is transductive, because we have the unlabeled test examples available. There is a substantial body of research on transductive classification \citep{joachims99a, sindhwani06, pechyony08}. The goal of transductive classification is to develop a classifier that will perform well on the entire test set.  Most transductive classification algorithms are inspired by either the large margin principle or the clustering principle. The large margin principle asserts that the decision boundary should correctly classify the training examples and pass through a low-density, sparse region of the test data. The clustering principle assumes that the classes form clusters, and that those clusters can be more accurately identified by including test points in the training process. 

In the transductive precision@$k$ problem, the goal is to select $k$ positive test points---the classifier or ranker is merely an intermediate step. Vapnik's principle \citep{vapnik98} dictates that we should not solve a more difficult problem on the way to solving the problem of interest. Hence, in this paper we jointly train the model and determine the threshold to obtain exactly $k$ predicted positive test instances. Our hypothesis is that the algorithm can take advantage of knowing the value of $k$ to choose a decision boundary that optimizes the precision of the $k$ predicted points. Note that this goal means that neither the large margin principle nor the clustering principle apply directly. A decision boundary that predicts exactly $k$ positives and has high precision on the training data is likely to pass through the region (or cluster) of positive instances rather than through a low-density region. Hence, the intuitions that underlie transductive classification do not provide guidance for solving the transductive precision@$k$ problem.

The paper proceeds as follows. We call the constraint that the model must predict exactly $k$ test instances as positives the {\it $k$-constraint}. We start by identifying a deterministic relation between the precision@$k$ measure and the accuracy of any classifier that satisfies the $k$-constraint. This suggests that the learning objective should maximize classifier accuracy subject to the $k$-constraint. We adopt the space of linear decision boundaries and introduce an algorithm we call Transductive optimization of Top $k$ precision (TTK). In the TTK optimization problem, the objective is to minimize the hinge loss on the training set subject to the $k$-constraint. This optimization problem is very challenging. It can be formulated a Mixed Integer Programming (MIP) problem. For small problems, the global optimum can be found by the branch-and-bound algorithm. To solve larger problems, we design a {\it feasible direction} method, which we find experimentally to converge very rapidly. An experiment comparing the feasible direction method to the exact MIP solution shows that solutions found by the feasible direction method are nearly optimal.  We also present a theoretical analysis of the transductive precision@$k$ problem which shows that one should train different scoring functions for different values of $k$.  

In the experiment section, we first present a small synthetic dataset to show how the TTK algorithm improves the SVM decision boundary. Then, we compare the TTK algorithm with five other algorithms on seven UCI datasets and three reserve design datasets. The results show that the TTK algorithm matches or exceeds the performance of these state-of-the-art algorithms on almost all of these datasets.



\section{The TTK model}
Let the distribution of the data be $\mathcal{D}$ with support in $\mathcal{X} \times \mathcal{Y}$.
In this work, we assume $\mathcal{X} = \mathcal{R}^d$ and only consider the binary classification problem with 
$\mathcal{Y} = \{-1, 1\}$. 
By sampling from $\mathcal{D}$ independently, a training set $(\bx, \by) = (x_i, y_i)_{i=1}^{n}$
and a test set $(\hbx, \hby) = (\hx_j, \hy_j)_{j=1}^{m}$ are obtained, but the 
labeling $\hby$ of the test set is unknown. The problem is to train a classifier and maximize the precision at $k$ on the test set. The hypothesis space is 
$\mathcal{H} \subset \mathcal{Y}^\mathcal{X}$ (functions mapping from $\mathcal{X}$ to $\mathcal{Y}$). 
The hypothesis $h \in \mathcal{H}$ is evaluated by the measure precision@$k$. 


When we seek the best classifier from $\mathcal{H}$ for selecting $k$ instances from the test set $\hbx$, we only consider classifiers satisfying the $k$-constraint, that is, these classifiers must be in the hypothesis space $\mathcal{H}_k(\hbx) = 
\{ h \in \mathcal{H} | \sum_{j=1}^{m} \indic[h(\hx_j) = 1] = k \}$, where $\indic[\cdot]$
is 1 if its argument is true and 0 otherwise.  All classifiers not predicting $k$ positives on the test set are excluded from $\mathcal{H}_k$. Note that with any two-step method with ranking and thresholding, the final classifier is in the hypothesis space $\mathcal{H}_k(\hbx)$.

\newcommand{\mtp}{m_\mathrm{tp}}
\newcommand{\mfp}{m_\mathrm{fp}}
\newcommand{\mtn}{m_\mathrm{tn}}
\newcommand{\mfn}{m_\mathrm{fn}}

To maximize the precision of $h \in \mathcal{H}_k(\hbx)$ on the test set, 
we essentially need to maximize the classification accuracy of $h$. This can be seen by the following relation. Let $m_{-}$ be the number of negative test instances, and let $\mtp$, $\mfp$ and $\mtn$ denote the number of true positives, false positives, and true negatives (respectively) on the test set as determined by $h$. Then the precision@$k$ of $h$ can be expressed as
\begin{eqnarray}
\label{prec_accuracy}
\rho(h) &=& \frac{1}{k} \mtp =\frac{1}{k} (\mtn + k - m_{-}) = \frac{1}{2k}(\mtp + \mtn + k - m_{-}).
\end{eqnarray}
Since the number of negative test instances $m_{-}$ is unknown but fixed, there is a deterministic relationship between the accuracy $(\mtp + \mtn) / m$ and the precision@$k$ on the test set. Hence, increasing classification accuracy directly increases the precision. This motivates us to maximize the accuracy of the classifier on the test set while respecting the $k$-constraint. 

In this section, we develop a learning algorithm for linear classifiers and thus 
$\mathcal{H} = \{h: \mathcal{X} \mapsto \mathcal{Y}, h(x; w, b) = \mbox{sign}(w^{\top} x + b)\}$. 
Our learning objective is to minimize the (regularized) hinge loss on the training set, which is a convex upper bound of the zero-one loss. Together with the $k$-constraint, the optimization problem is
\begin{align}
\min_{w, b} & ~ \frac{1}{2} \|w\|_2^2 + C \sum_{i=1}^{n} [1 - y_i~(w^{\top}x_i + b)]_+, 
\label{opt}\\
s.t. \hspace{1cm} &    \sum_{j = 1}^{m} \mathcal{I}[w^\top \hx_j + b > 0] = k ~, \nonumber
\end{align}
where $[\cdot]_+ = \max(\cdot, 0)$ calculates the hinge loss on each instance. Due to the piece-wise constant function in the constraint, the problem is very hard to solve. 

Let us relax the equality constraint to an inequality constraint. The optimization problem becomes 
\begin{align}
\min_{w, b} & ~ \frac{1}{2} \|w\|_2^2 + C \sum_{i=1}^{n} [1 - y_i~(w^{\top}x_i + b)]_+, 
\label{opt_relax}\\
s.t. \hspace{1cm} &    \sum_{j = 1}^{m} \mathcal{I}[w^\top \hx_j + b > 0] \le k ~.\nonumber
\end{align}
This relaxation generally does not change the solution to the optimization problem. If we neglect the constraint, then the solution that minimizes the objective will be an SVM. In our applications, there are typically significantly more than $k$ positive test points, so the SVM will usually predict more than $k$ positives. In that case, the inequality constraint will be active, and the relaxed optimization problem will give the same solution as the original problem \footnote{In the extreme case that many data points are nearly identical, the original problem may not have a solution, and the relaxed constraint is satisfied by the ``less than'' relation. This is unlikely to arise in practice.}. 

Even with the relaxed constraint, the problem is still hard, because the feasible region is non-convex. We first express the problem as a Mixed Integer Program (MIP). Let $G$ be a large constant and $\eta$ be a binary vector of length $m$. Then we can write the optimization problem as 
\begin{align} 
\min_{w, b, \eta} & ~ \frac{1}{2} \|w\|_2^2 + C \sum_{i=1}^{n} [1 - y_i~(w^{\top}x_i + b)]_+, 
\label{opt_mip}\\
s.t. \hspace{1cm} &    w^\top \hx_j + b \le \eta_{j} G, ~~ j = 1, \ldots, m  \nonumber\\
& \eta_j \in \{0, 1\}, ~~ j = 1, \ldots, m  \nonumber \\
& \sum_{j = 1}^{m} \eta_j = k \nonumber
\end{align}
The equivalence of (\ref{opt_relax}) and (\ref{opt_mip}) is easy to show, and we omit the proof here. For this MIP, a globally optimal solution can be found for small problem instances via the branch-and-bound method with an off-the-shelf package. We used Gurobi\cite{gurobi15}. 

For large problem instances, finding the global optimum of the MIP is impractical. We propose to employ a {\it feasible direction} algorithm \citep{bazaraa06}, which is an iterative algorithm designed for constrained optimization. In each iteration, it first finds a descending direction in the feasible direction cone and then calculates a step size to make a descending step that leads to a improved feasible solution. The feasible direction algorithm fits this problem well. Because the constraint is a polyhedral cone, a step in any direction within the feasible direction cone will generate a feasible solution provided that the step length is sufficiently small. Since our objective is convex but the constraint is highly non-convex, we want to avoid making descending steps along the constraint boundary in order to take bigger steps and avoid local minima.  

In each iteration, we first need to find a descending direction. The subgradient $(\nabla w, \nabla b)$ of the objective with respect to $(w, b)$ is calculated as follows. Let $\xi_i = 1 - y_i~(w^{\top}x_i + b)$ be the hinge loss on instance $i$. Then 
\begin{eqnarray}
\nabla w = w - C \sum_{i: \xi_i > 0 }  y_i x_i ~, ~~~~ \nabla b = - C \sum_{i: \xi_i  > 0 }  y_i. 
\end{eqnarray}
We need to project the negative subgradient $(- \nabla w, - \nabla b)$ to a feasible direction to get a feasible descending direction. Let $L$, $E$, and $R$ be the sets of test instances predicted to be positive, predicted to be exactly on the decision boundary, and predicted to be negative: 
\begin{eqnarray}
L = \{j : ~~w^{\top} \hx_j + b > 0\}, ~~~~ E = \{j: ~~w^{\top} \hx_j + b = 0\}, ~~~~ R = \{j: ~~w^{\top} \hx_j + b < 0\}. \nonumber
\end{eqnarray}

With the $k$-constraint, the feasible direction cone can be written as 
\begin{eqnarray}
\mathcal{F} = \left\{ (d_w, d_b) : \sum_{j \in E} \mathcal{I}[\hx_j^\top d_w + d_b > 0] + |L| \le k\right\}.
\end{eqnarray}

We do not project the negative gradient onto $\mathcal{F}$, both because this is computationally difficult and because it often gives a direction along the boundary of $\mathcal{F}$.  Instead, we find a descending direction by  projecting the negative gradient into the null space of a set $B \subseteq E$ of test instances. We first sort the instances in $E$ in descending order according to the value of $- \hx_j^\top \nabla w - \nabla b$. Let $j': 1 \le j' \le |E|$  re-index the instances in this order. To construct the set $B$, we start with $B = \emptyset$ and the initial direction $(d_w, d_b) = - (\nabla w, \nabla b)$. The starting index is $j_0 = 1$ if $|L| = k$, and $j_0 = 2$ if $|L| < k$. Then with index $j'$ starting from $j_0$ and increasing, we consecutively put instance $j'$ into $B$ and project $(d_w, d_b)$ into the null space of $\{(\hx_{j^\circ}, 1) : j^\circ \in B\}$. We stop at $j' = j_1$ when all the remaining instances in $E$ have negative inner product\footnote{By ``inner product'' between a direction $(d_w, d_b)$ and an instance $x$, we mean $x^\top d_w + d_b$.} with $(d_w, d_b)$. The final projected direction is denoted by $(d_w^\star, d_b^\star)$. The direction $(d_w^\star, d_b^\star)$ has non-positive inner product with all instances with indices from $j'= j_0$ to $j' = |E|$, so these instances will not move into the set $L$ when $(w, b)$ moves in that direction. Only when $|L| < k$, is the first instance allowed to move from $E$ to $L$. It is easy to check that the final projected direction $(d_w^\star, d_b^\star)$ is in the feasible cone $\mathcal{F}$ and that it is a descending direction. This subgradient projection algorithm is summarized in Algorithm \ref{alg}.

\begin{algorithm}[tb]
   \caption{Find a descending feasible direction}
   \label{alg}
\begin{algorithmic}
   \STATE {\bfseries Input:} subgradient $(\nabla w, \nabla b)$, instance set $\{\hx_j: j \in E \}$, size $|L|$, $k$
   \STATE {\bfseries Output:} descending feasible direction $(d_w^\star, d_b^\star)$
   \STATE Sort instances in $E$ in descending order according to $- \hx_{j}^\top \nabla w - \nabla b$
   \STATE Initialize $(d_w, d_b) = - (\nabla w, \nabla b)$
   \STATE Initialize $B = \emptyset$
   \STATE $j_0 = \min(k - |L|, 1) + 1$
   \FOR{$j' = j_0$ {\bfseries to} $|E|$}
   \IF{ $\exists j'': \; j' \le j'' \le |E|, ~ \hx_{j''}^\top d_w + d_b > 0$ } 
   \STATE $B = B \cup \{j'\}$
   \STATE project $(d_w, d_b)$ into the null space of $\{(\hx_{j^\circ}, 1) : j^\circ \in B \}$ 
   \ELSE
   \STATE {\bf break} 
   \ENDIF
   \ENDFOR
   \STATE $d_w^\star = d_w$, $d_b^\star = d_b$
\end{algorithmic}
\end{algorithm}

In this design, we have the following considerations. When $|L| < k$, the instance that has the largest inner product with the negative subgradient is not used to constrain the projected direction, since this can avoid a very hard constraint on the projection. We allow at most one instance to move from $E$ to $L$ to reduce the chance that $(w, b)$ hits the boundary. In the projecting iterations, instances with large inner products are selected first to reduce the number of projections. 

Once a descending direction is chosen, we perform a line search to determine the step size. We first find the maximum step size $\alpha$ that guarantees the feasibility of the descending step. That is, no points in $R$ will cross the decision boundary and enter $S$ with the step length $\alpha$.
\begin{eqnarray}
\alpha = \min_{j \in R ~:~ \hx_j^\top d_w^\star + d_b > 0} ~~\frac{ - (\hx_j^\top w + b)}{\hx_j^\top d_w^\star + d_b}. \label{steplength}
\end{eqnarray}


Then we do a line search in $[0, 0.5\alpha]$ to find the best step length $\alpha^\star$. Note that the objective function is a convex piece-wise quadratic function, so we only need to check these elbow points plus a minimum between two elbow points to find the best step length. We omit the details. The shrinkage $0.5$ of $\alpha$ reduces the chance of $(w, b)$ hitting the boundary. 

We initialize $w$ by training a standard linear SVM (although any linear model can be used) and then initialize $b$ to satisfy the $k$-positive constraint. This gives us a pair $(w, b)$ that is a feasible solution to (\ref{opt_relax}). Then $(w, b)$ is updated in each iteration according to $(w, b) := (w, b) + \alpha^\star (d_w^\star, d_b^\star)$ until convergence. 

We set the maximum number of iterations, $T$, to 500; the algorithm typically requires only 200-300 iterations to converge. In each iteration, the two most expensive calculations are computing the subgradient and projecting the negative subgradient. The first calculation requires $O(nd)$ operations, and the second one takes $O(d^3)$ operations, since there are usually no more than $(d+1)$ instances in the set $E$. The overall running time is the time of training an initial model plus $O(T(nd + d^3))$. 

Though motivated differently, the AATP algorithm solves a similar optimization problem. The AATP objective is equivalent to ours---but applied to the training set. Their constraint is that the top $q$ quantile of training instances must receive positive scores and all others, negative scores. The AATP authors assume that the decision boundary must go though a training instance, so their relaxation of the optimization problem is constrained to require one instance to be on the decision boundary, classify {\it at least} quantile $q$ training examples as positive, and minimize the training set hinge loss. This tends to find solutions that classify more than quantile $q$ of the instances as positive. We will see below that when we compute the exact optimum using an MIP solver, that solution often has many instances on the decision boundary.  This demonstrates that the AATP relaxation is quite loose.

\section{Analysis}
Before presenting experiments, we first argue that different values of $k$ require us, in general, to train different models. We work with the population distribution  $\mathcal{D}$ instead of with samples, and we assume linear models. Suppose the distributions of positive instances and negative instances have probability measures $\mu_+$ and $\mu_-$ defined on $\mathcal{R}^d$. The total distribution is a mixture of the two distributions, and it has  measure $\mu =  \lambda \mu_+ + (1 - \lambda)\mu_{-}$ with $\lambda \in (0, 1)$. The classifier $(w, b)$ defines a positive region $R_{w, b} = \{x \in \mathcal{R}^d, w^{\top}x + b > 0\}$. Assume $\mu_+(R_{w, b})$ and $\mu_-(R_{w, b})$ are both differentiable with respect to $(w, b)$. If we consider classifiers that classify fraction $q$  of the instances as positive, then $\mu(R_{w, b}) = q$. The precision of the classifier will be $ \lambda \mu_+(R_{w, b}) ~/~ q$. The optimal classifier is therefore
\begin{align}
(w^\star, ~ b^\star) ~~~~~ = &  ~~~~~ \arg \max_{(w, b)}  ~~~~~\lambda \mu_+(R_{w, b})  \label{eq-pop-opt}\\
&  s.t.  ~~~~~ \lambda \mu_+(R_{w, b}) + (1 - \lambda) \mu_-(R_{w, b}) = q. \nonumber
\end{align}

If we change $q$, we might hope that we do not need to modify $w^\star$ but instead can just change $b^\star$. However, this is unlikely to work. 
\begin{theorem}
If $(w^\star, b_1)$ and $(w^\star, b_2)$ are two optimal solutions for \eqref{eq-pop-opt} with two different quantile values $q_1$ and $q_2$, then
\begin{eqnarray} 
\exists s_1, t_1, s_2, t_2, \in \mathbf{R}, && s_1 \frac{\partial \mu_+ (R_{w^\star, b_1})}{\partial (w^\star, b_1)} = t_1 \frac{\partial \mu_- (R_{w^\star, b_1})}{\partial (w^\star, b_1)}, \label{vec_eq1} \\
&& s_2 \frac{\partial \mu_+ (R_{w^\star, b_2})}{\partial (w^\star, b_2)} = t_2 \frac{\partial \mu_- (R_{w^\star, b_2})}{\partial (w^\star, b_2)}. \label{vec_eq2} 
\end{eqnarray}
\end{theorem}
The proof follows directly from the KKT conditions.
Note that \eqref{vec_eq1} and  \eqref{vec_eq2} are two vector equations. When $b_1$ is changed into be $b_2$, the vectors of partial derivatives, ${\partial \mu_+ (R_{w^\star, b_1})} / {\partial (w^\star, b_1)}$ and ${\partial \mu_- (R_{w^\star, b_1})}/ {\partial (w^\star, b_1)}$ must change their directions in the same way to maintain optimality. This will only be possible for very special choices of $\mu_+$ and $\mu_-$. This suggests that $(w^\star, b^\star)$ should be optimized jointly to achieve each target quantile value $q$.


\section{Experimental Tests}

\newcommand{\wsvm}{w_\mathrm{svm}}
\newcommand{\waatp}{w_\mathrm{aatp}}
\newcommand{\wttk}{w_\mathrm{ttk}}
\newcommand{\wttkm}{w_\mathrm{ttk}^\star}

\newcommand{\ttkmip}{TTK$_\mathrm{MIP}$}
\newcommand{\ttkfd}{TTK$_\mathrm{FD}$}

\subsection{An illustrative synthetic dataset}

We begin with a simple synthetic example to provide some intuition for how the TTK algorithm improves the SVM decision boundary, see Figure \ref{synthetic}. The dataset consists of 40 training and 40 test instances. The training and testing sets each contain 22 positive and 18 negative instances.  Our goal is to select $k=4$ positive test instances. The bold line is the decision boundary of the SVM. It is an optimal linear classifier both for overall accuracy and for precision@$k$ for $k=24$. However, when we threshold the SVM score to select 4 test instances, this translates the decision boundary to the dashed line, which gives very poor precision of 0.5. This dashed line is the starting point of the TTK algorithm. After making feasible direction descent steps, TTK finds the solution shown by the dot-dash-dot line. The $k$ test instances selected by this boundary are all positive.  Notice that if $k=24$, then the SVM decision boundary gives the optimal solution. This provides additional intuition for why the TTK algorithm should be rerun whenever we change the desired value of $k$.

\subsection{Effectiveness of Optimization}

\begin{figure}[t]
\begin{minipage}{0.47\textwidth}
\centerline{\includegraphics[width=0.8\columnwidth,trim={2.7cm 5.5cm 1cm 5cm},clip]{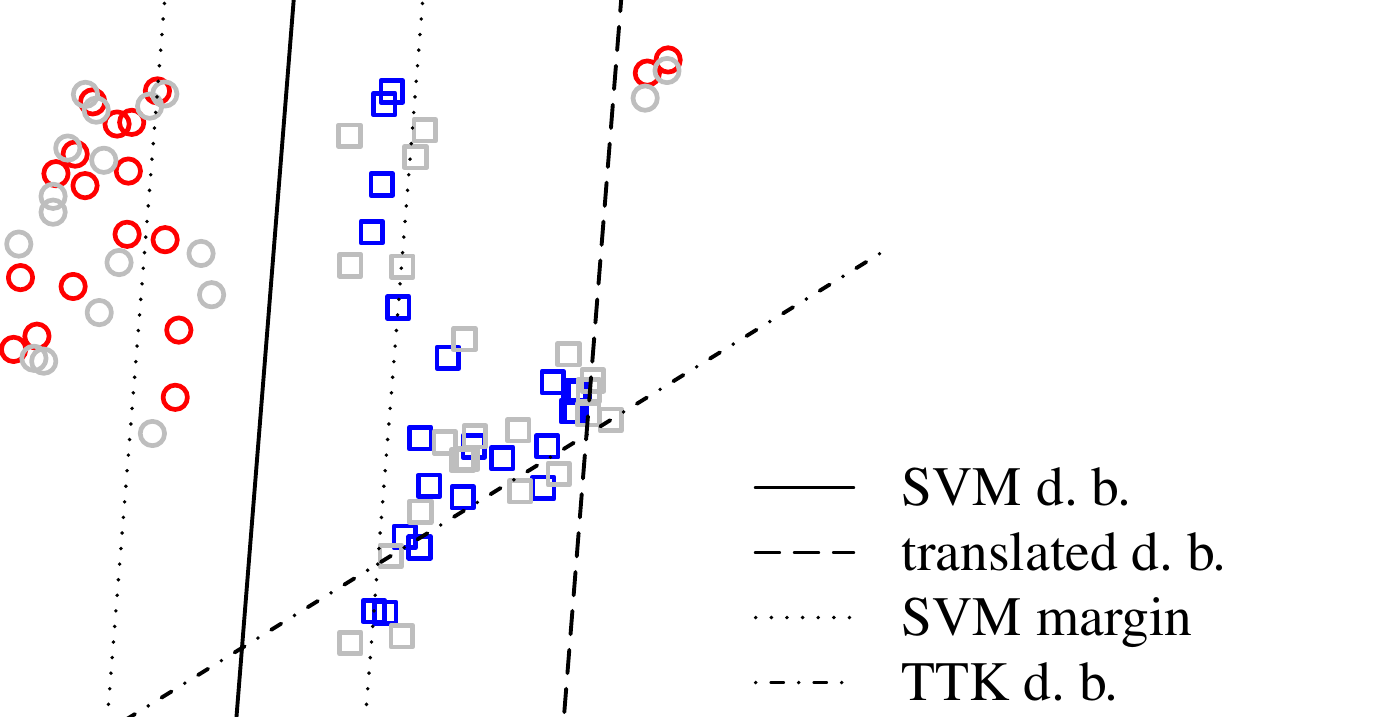}}
\caption{TTK improves the SVM decision boundary (``d. b.''). Square/circle: positive/negative instance, colored/gray: training/testing instance. $k$ = 4.}
\label{synthetic}
\vskip -0.2in
\end{minipage} \hfill
\begin{minipage}{0.52\textwidth}
\captionof{table}{Training and test loss attained by different methods}
\label{loss_table}
\vspace*{0.1in}
\scalebox{0.8}{
\begin{tabular}{c|cccc}
\hline 
dataset & \small SVM  & \small AATP & \small \ttkmip & \small \ttkfd \\
\hline
diabetes \tabularnewline
train obj. & 311 $\pm$ 25& 265 $\pm$ 23& 224 $\pm$ 7& 226 $\pm$ 7\tabularnewline
test loss & 323 $\pm$ 20& 273 $\pm$ 24& 235 $\pm$ 6& 235 $\pm$ 5\tabularnewline
\hline
ionosphere \tabularnewline
train obj. & 325 $\pm$ 46& 474 $\pm$ 88& 127 $\pm$ 4& 136 $\pm$ 4\tabularnewline
 test loss & 338 $\pm$ 44& 488 $\pm$ 85& 146 $\pm$ 5& 150 $\pm$ 7\tabularnewline
\hline
sonar \tabularnewline
train obj. & 167 $\pm$ 52& 166 $\pm$ 41& 20 $\pm$ 8& 30 $\pm$ 10\tabularnewline
test loss & 216 $\pm$ 22& 213 $\pm$ 30& 103 $\pm$ 19& 105 $\pm$ 24\tabularnewline
\hline
\end{tabular}
}
\end{minipage}
\end{figure}

One way to compare different algorithms is to see how well they optimize the training and test surrogate loss functions. We trained a standard SVM, AATP, TTK (MIP) and TTK (feasible direction) on three UCI\footnote{https://archive.ics.uci.edu/ml/datasets.html} data sets: {\it diabetes}, {\it ionosphere } and {\it sonar}. For the SVM and AATP methods, we fit them to the training data and then obtain a top-$k$ prediction by adjusting the intercept term $b$. We set $k$ to select 5\% of the test instances. Table~\ref{loss_table} reports the regularized hinge loss on the training set and the hinge loss on the test set. The hyper-parameter $C$ is set to 1 for all methods. The results show that TTK with either solver obtains much lower losses than the competing methods. From the difference between the third (MIP) and the fourth (feasible direction) columns, we can also see that the feasible direction method finds near-optimal solutions. 

To understand and compare the behavior of AATP and TTK, we performed a non-transductive experiment (by making the training and test sets identical). We measured the number of training instances that fall on the decision boundary and the fraction of training instances classified as positive (see  Table \ref{train_info}). The optimal solution given by the MIP solver always puts multiple instances on the decision boundary, whereas the AATP method always puts a single instance on the boundary.  The MIP always exactly achieves the desired $k$, whereas AATP always classifies many more than $k$ instances as positive. This shows that the AATP assumption that the decision boundary should pass through exactly one training instance is wrong.

\begin{table}[t]
\centering
\caption{AATP and TTK solution statistics: Number of instances on the decision boundary ("\# at d.b.") and fraction of instances predicted as positive ("fraction +")}
\label{train_info}
\vspace*{0.1in}
\scalebox{0.9}{
\begin{tabular}{r|cc|cc}
\hline
\multirow{2}{*}{dataset, dimension, ratio of positives} & \multicolumn{2}{c}{\small AATP} & \multicolumn{2}{c}{\small \ttkmip} \\
& \# at d.b. & fraction +  & \# at d.b. & fraction +\\
\hline 
diabetes, $d=8$, $n_+/n = 0.35$    & 1 & 0.12 & 5  & 0.05 \\
ionosphere, $d=33$, $n_+/n = 0.64$ & 1 & 0.53 & 21 & 0.05 \\
sonar, $d=60$, $n_+/n = 0.47$      & 1 & 0.46 & 40 & 0.05 \\
\hline
\end{tabular}}
\end{table}

\subsection{Precision evaluation on real-world datasets}

In this subsection, we evaluate our TTK method on ten datasets. Seven datasets 
\{{\it diabetes, ionosphere, sonar, spambase, splice}\} 
from UCI repository and 
\{{\it german-numer, svmguide3}\} from the LIBSVM web site\footnote{http://www.csie.ntu.edu.tw/~cjlin/libsvmtools/datasets/binary.html}, are widely studied binary classification datasets. The other three datasets, NY16, NY18 and NY88, are three species distribution datasets extracted from a large eBird dataset \cite{sullivan09}; each of them has 634 instances and 38 features. The eBird dataset contains a large number of checklists of bird counts reported from birders around the world. Each checklist is associated with the latitude and longitude of the observation and a set of 38 features describing the habitat. We chose a subset of the data consisting of checklists of three species from New York state in June of 2012. To correct for spatial sampling bias, we formed spatial cells by imposing a grid over New York and combining all checklists reported within each grid cell. This gives 634 cells (instances). Each instance is labeled with whether a species was present or absent in the corresponding cell. 

We compare the TTK algorithm with 5 other algorithms. The SVM algorithm \cite{scholkopf2002} is the baseline. The Transductive SVM \cite{joachims99a} compared here (denoted by TSVM) uses the UniverSVM  \cite{sinz12} implementation, which optimizes its objective with the convex-concave procedure. SVMperf \cite{joachims05} can optimize multiple ranking measures and it is parameterized here to optimize precision@$k$. Two algorithms, Accuracy At The Top (AATP) \cite{boyd12} and TopPush \cite{li14},  are specially designed for top precision optimization. 
The proposed TTK objective is solved using both the MIP solver and the  feasible direction method (denoted \ttkmip and \ttkfd, respectively). 
Each algorithm is run 10 times on 10 random splits of each dataset. Each of these algorithms requires setting the regularization parameter $C$. This was done by performing five 2-fold internal cross-validation runs within each training set and selecting the  value of $C$ from the set $\{0.01, 0.1, 1, 10, 100\}$ that maximized precision on the top 5\% of the (cross-validation) test points. With the chosen value of $C$, the algorithm was then run on the full training set (and unlabeled test set) and the precision on the top 5\% was measured. The achieved precision values were then averaged across the 10 independent runs. 

Table \ref{perf1} shows the performance of the algorithms. For datasets with more than 1000 instances, the AATP and \ttkmip algorithms do not finish within a practical amount of time, so results are not reported for these algorithms on those datasets. This is indicated in the table by ``NA''. The results for each pair of algorithms are compared by a paired-differences t-test at the $p<0.05$ significance level. If one algorithm is not significantly worse than any of the other algorithms, then it is regarded as one the best and its performance is shown in bold face. Wins, ties and losses of of \ttkmip and \ttkfd with respect to all other algorithms are reported in the last two rows of Table \ref{perf1}.

On each of the six small datasets, the performance of \ttkmip ~ matches or exceeds that of the other algorithms.  The \ttkfd method does almost as well---it is among the best algorithms on 8 of the 10 datasets. It loses once to SVMperf (on svmguide3) and once to AATP (on ionosphere). None of the other methods performs as well.  By comparing \ttkfd with SVM, we see that the performance is improved on almost all datasets, so the \ttkfd method can be viewed as a safe treatment of the SVM solution. As expected, the transductive SVM does not gain much advantage from the availability of the testing instances, because it seeks to optimize accuracy rather than precision@$k$. The TopPush algorithm is good at optimizing the precision of the very top instance. But when more positive instances are needed, the TopPush algorithm does not perform as well as TTK. 

\begin{table*}[t]
\centering
\caption{Mean Precision ($\pm$ 1 standard deviation) of classifiers when \%5 of testing instances are predicted as positives.}
\label{perf1}
\scalebox{0.94}{
\begin{tabular}{c||ccccc|cc}
\hline 
dataset & \small SVM & \small TSVM &\small SVMperf &\small TopPush &  \small AATP 
& {\small TTK$_{\mathrm{MIP}}$} & {\small TTK$_{\mathrm{FD}}$} \\
\hline 
diabetes & {\bf .86$\pm$.08 }& {\bf .86$\pm$.09 }& .69$\pm$.20 & {\bf .80$\pm$.10 }& .68$\pm$.28 & {\bf .85$\pm$.10 }& {\bf .86$\pm$.08 }\tabularnewline
ionosphere & .76$\pm$.13 & .80$\pm$.17 & .82$\pm$.22 & .71$\pm$.16 & {\bf 1.00$\pm$.00 }& {\bf .97$\pm$.05 }& .84$\pm$.15 \tabularnewline
sonar & {\bf .96$\pm$.08 }& {\bf .98$\pm$.06 }& .85$\pm$.16 & .88$\pm$.13 & .90$\pm$.11 & {\bf .96$\pm$.08 }& {\bf 1.00$\pm$.00 }\tabularnewline
german-numer & .70$\pm$.08 & {\bf .72$\pm$.08 }& {\bf .56$\pm$.17 }& .63$\pm$.12 & NA. & NA. & {\bf .71$\pm$.06 }\tabularnewline
splice & {\bf 1.00$\pm$.00 }& {\bf 1.00$\pm$.00 }& {\bf 1.00$\pm$.01 }& {\bf 1.00$\pm$.00 }& NA. & NA. & {\bf 1.00$\pm$.00 }\tabularnewline
spambase & .97$\pm$.02 & .97$\pm$.02 & {\bf .98$\pm$.01 }& .96$\pm$.02 & NA. & NA. & {\bf .98$\pm$.01 }\tabularnewline
svmguide3 & .86$\pm$.07 & .85$\pm$.07 & {\bf .91$\pm$.04 }& .83$\pm$.07 & NA. & NA. & .87$\pm$.06 \tabularnewline
NY16 & .64$\pm$.08 & .64$\pm$.09 & {\bf .65$\pm$.12 }& .62$\pm$.10 & .62$\pm$.08 & {\bf .68$\pm$.07 }& {\bf .70$\pm$.09 }\tabularnewline
NY18 & {\bf .44$\pm$.11 }& {\bf .45$\pm$.10 }& .36$\pm$.07 & {\bf .43$\pm$.13 }& {\bf .46$\pm$.12 }& {\bf .46$\pm$.08 }& {\bf .47$\pm$.12 }\tabularnewline
NY88 & {\bf .40$\pm$.08 }& .33$\pm$.12 & {\bf .37$\pm$.15 }& .34$\pm$.08 & .31$\pm$.09 & {\bf .40$\pm$.09 }& {\bf .42$\pm$.07 }\tabularnewline
\hline
{\small TTK$_{\mathrm{MIP}}$ } w/t/l &  1/5/0  &  2/4/0  &  2/4/0  &  1/5/0  &  2/4/0  &  & \tabularnewline
{\small TTK$_{\mathrm{FD}}$} w/t/l &  3/7/0  &  4/6/0  &  3/6/1  &  7/3/0  &  4/1/1  &  & \tabularnewline
\hline

\hline
\end{tabular}
}
\end{table*}

\section{Summary}
This paper introduced and studied the transductive precision@$k$ problem, which 
is to train a model on a labeled training set and an unlabeled test set
and then select a fixed number $k$ of positive instances from the testing set. Most existing methods first train a scoring function and then adjust a threshold to select the top $k$ test instances. We show that by learning the scoring function and the threshold together, we are able to achieve better results. 

We presented the TTK method. The TTK objective is the same as the SVM objective, but TTK imposes the constraint that the learned model must select exactly $k$ positive instances from the testing set. This constraint guarantees that the final classifier is optimized for its target task. The optimization problem is very challenging, since it involves a set selection problem. We introduced two algorithms for solving it. First, we formulated it as a mixed integer program and solved it exactly via the branch-and-bound method. Second, we designed a feasible direction algorithm that is able to scale to larger datasets but that attempts to find a good solution. We compared both TTK algorithms to several state-of-the-art methods on ten datasets. The results indicate that the performance of the TTK methods matches or exceeds all of the other algorithms on most of these datasets.  

Our analysis and experimental results show that the TTK objective is a step in the right direction. However, we believe that the performance can be further improved if we can minimize a tighter (possibly non-convex) bound on the zero-one loss. In the future, we will extend the TTK formulation to problems with multiple labels so that it can be applied to reserve design problems involving multiple species. 

\newpage


\small
{
\bibliography{trans_prec}
}

\end{document}